\pdfoutput=1

\documentclass[11pt]{article}

\usepackage{EMNLP2023}

\usepackage{times}
\usepackage{latexsym}

\usepackage[T1]{fontenc}

\usepackage[utf8]{inputenc}

\usepackage{microtype}
\usepackage{booktabs}
\usepackage{paralist}
\usepackage{csquotes}
\usepackage{multirow}
\usepackage{amsmath}
\usepackage{makecell}
\usepackage{graphicx}

\usepackage{inconsolata}

\title{Outcome-Constrained Large Language Models for Countering Hate Speech}

\author{Lingzi Hong \\
	University of North Texas \\
	\texttt{lingzi.hong@unt.edu} \\\And
	Pengcheng Luo \\
	Peking University\\
	\texttt{luopc@pku.edu.cn} \\\AND
	Eduardo Blanco \\
	University of Arizona\\
	\texttt{eduardoblanco@arizona.edu} \\\And
	Xiaoying Song \\
	University of North Texas\\
	\texttt{XiaoyingSong@my.unt.edu} \\}

\begin{document}
\maketitle
\begin{abstract}
Automatic counterspeech generation methods have been developed to assist efforts in combating hate speech. 
Existing research focuses on generating counterspeech with linguistic attributes such as being polite, informative, and intent-driven. However, the real impact of counterspeech in online environments is seldom considered. 
This study aims to develop methods for generating counterspeech constrained by conversation outcomes and evaluate their effectiveness. 
We experiment with large language models (LLMs) to incorporate into the text generation process two desired conversation outcomes:
low conversation incivility and non-hateful hater reentry.
Specifically, we experiment with \textit{instruction prompts}, \textit{LLM finetuning}, and \textit{LLM reinforcement learning (RL)}.
Evaluation results show that our methods effectively steer the generation of counterspeech towards the desired outcomes.
Our analyses, however, show that 
there are differences in the quality and style depending on the model.
\end{abstract}

\section{Introduction}

Hate speech has posed significant challenges to healthy and productive online communication. 
Counterspeech, which involves using constructive, positive, or factual responses to challenge or counteract hate speech, has shown to be effective in moderating online hostilities~\cite{buerger2021iamhere}, promoting productive user engagement~\cite{mivskolci2020countering}, and educating online users ~\cite{blaya2019cyberhate}.

Automatic generation of counterspeech has been researched to support timely and effective efforts to fight hate speech. Synthetic counterspeech datasets have been developed using crowdsourcing~\cite{qian2019benchmark} and human-in-the-loop strategies~\cite{chung2021towards}.
These datasets have been used to develop counterspeech generation models. 
However, the impact of counterspeech in online environments has not been considered in the dataset creation.
As a result, it is unknown whether generated counterspeech elicits civil or hateful follow-up conversations.

Recent counterspeech generation research focused on constrained generation with linguistic attributes (e.g., being polite, emotion-laden~\cite {saha2022countergedi}), or embedded with knowledge~\cite{chung2021towards}.
Questions about the impact of counterspeech with such attributes linger. Previous research also found one of the barriers counterspeakers face is their inability to determine the potential impact of counterspeech~\cite{mun2024counterspeakers}.
However, there is a lack of research on generating outcome-oriented counterspeech, e.g., speech that leads to desired outcomes such as de-escalating user conflicts or encouraging constructive engagement in follow-up conversations. 

\begin{table*}[ht!]
	\centering
	\small
	\begin{tabular}{p{3cm}  p{3.2cm} p{3.5cm} p{4.4cm}}
		\toprule
		Prior Work                 & Constraint  & Hate Speech & Generation Method  \\
		\toprule
		CONAN \newline\cite{chung2019conan} &  None & Islamophobic & Expert-based and LM data augmentation\\
		\addlinespace
		Benchmark \newline \cite{qian2019benchmark} & None & Reddit, Gab & Crowdsourcing and LM generation \\
		\addlinespace
		MultiCONAN \newline \cite{fanton2021human} & None & Multiple hate targets & LLM generation with review/edits by experts \\
		\addlinespace
		Knowledge \newline \cite{chung2021towards} & Informative & CONAN & LLM generation with information from knowledge repository \\
		\addlinespace
		Generate-Prune \newline \cite{zhu2021generate} & Diverse and relevant & Benchmark, CONAN & LLM generation with quality classifier \\
		\addlinespace
		COUNTERGEDI \newline \cite{saha2022countergedi} & Polite, detoxified, and emotional & Benchmark, CONAN & DialoGPT and GEDI for constraint generation \\
		\addlinespace
		Intent \newline \cite{gupta2023counterspeeches} & Multiple intents & CONAN,  MultiCONAN & QUARC with intent category representation and fusion \\
		\midrule
		Ours & Expected outcomes & Benchmark, CONAN, MultiCONAN & LLMs: instruction prompting, finetuning, and  RL \\
		\bottomrule
	\end{tabular}
	\caption{Summary of recent work on counterspeech generation, including dataset creation and modeling efforts.}
	\label{t:relatedwork}
\end{table*}

Notably, previous studies indicate that language may influence the development of a conversation,
including 
discourse popularity~\cite{horawalavithana2022online},
reentry behaviors~\cite{wang2021re},
and the rise of hate speech~\cite{liu2018forecasting}.
This leads to our research questions:
\begin{compactitem}
	\item How can constraints on conversation outcomes be incorporated into developing LLMs for generating counterspeech?
	\item How effective are these methods in generating outcome-oriented counterspeech?
\end{compactitem}

Unlike previous work that considers explicit linguistic attributes to guide language generation,
we formulate counterspeech generation to achieve desired outcomes (e.g., constructive user engagement). 
Our study holds potential for broader applications. Anticipating the direction of a conversation is crucial in crafting effective responses, allowing the conversation to meet the objectives (e.g., reducing hate speech, altering user behavior, and promoting positive discourse). 
This study makes the following contributions: 
(i) introducing conversation outcomes as a constraint to guide the generation of counterspeech, 
(ii) experimenting with LLMs for generating outcome-constrained counterspeech using \textit{instruction prompts}, \textit{LLM finetuning}, and \textit{LLM reinforcement learning (RL)},
and 
(iii) evaluating counterspeech generation models with various metrics to understand the strengths and weaknesses of the methods. 

\section{Related Work}

\paragraph{Generating Counterspeech}
Table~\ref{t:relatedwork} presents recent work on counterspeech generation.
CONAN has counterspeech written by NGO experts and augmented by language models~\cite{chung2019conan};
Benchmark was built with hate speech from Gab and Reddit and counterspeech created by crowdsourcing workers~\cite{qian2019benchmark};
and
MultiCONAN is a high-quality, high-quantity dataset created by experts coupled with language model generation for hate speech with multiple targets~\cite{fanton2021human}. 
Counterspeech generation models have been built with these datasets~%
\cite{halim2023wokegpt,tekirouglu2020generating,tekirouglu2022using,bonaldi2024nlp}.
Unlike us, none consider conversation outcomes elicited by the generated counterspeech.

Researchers have investigated counterspeech generation under constraints.
\citet{chung2021towards} proposed a generation pipeline grounded in external knowledge repositories to generate more informative and less biased replies.
\citet{zhu2021generate} proposed to generate more diverse and relevant counterspeech by developing a three-stage pipeline that uses LLMs to generate candidates, prunes the ungrammatical ones, and selects the best instances. 
\citet{saha2022countergedi} proposed an ensemble generative discriminator to generate more polite, detoxified, and emotion-laden counterspeech. 
\citeauthor{gupta2023counterspeeches} (\citeyear{gupta2023counterspeeches}) developed IntentCONAN, where the generation of counterspeech is conditioned on five intents: informative, denouncing, questioning, positive, and humorous. 
Similarly, \citet{fraser2023makes}  utilized ChatGPT to generate counter-stereotype text by incorporating countering strategies in queries. \citet{hassan2023discgen} proposed prompting strategies based on discourse theories to generate more context-relevant counterspeech. 
There are also studies on the generation of counterspeech in languages other than English (e.g., Italian~\cite{chung2020italian}).
Unlike us, none of these previous works generate counterspeech to elicit positive behaviors in the follow-up conversations.

\paragraph{Language Generation with Constraints}
Extensive studies have targeted language generation under complex lexical constraints
such as formality~\cite{jin2022deep}, text with certain concepts~\cite{lu2022neurologic}, dialogue that takes latent variables~\cite{bao2020plato}, and knowledge-enhanced text~\cite{yu2022survey}.
Not all styles can be described explicitly as linguistic attributes.
Indeed, some `styles' can only be defined in a data-driven way based on the shared attributes across various datasets~\cite{mou2020stylized}. 
In this study, we generate counterspeech very likely to lead to desired conversational outcomes. 

Methods have been developed for constrained language generation. 
\citet{wang2018sentigan} proposed the SentiGAN framework to generate text with a given sentiment.
\citet{kumar2021controlled} proposed MUCOCO to allow for controllable inference with multiple attributes as constraints to the optimization. 
\citet{krause2021gedi} developed GeDi, a discriminator-based approach to guide the decoding process in language generation. It enables text generation with desired or undesired attributes. 
\citet{schick2021self} proposed a self-debiasing approach to reduce the probability of language models generating problematic text.
Unlike these previous efforts, we experiment with methods to adjust language model-generated texts to achieve specific conversational outcomes.


\section{Methodology}

\begin{figure}[t]
	\centering
	\includegraphics[width=\columnwidth]{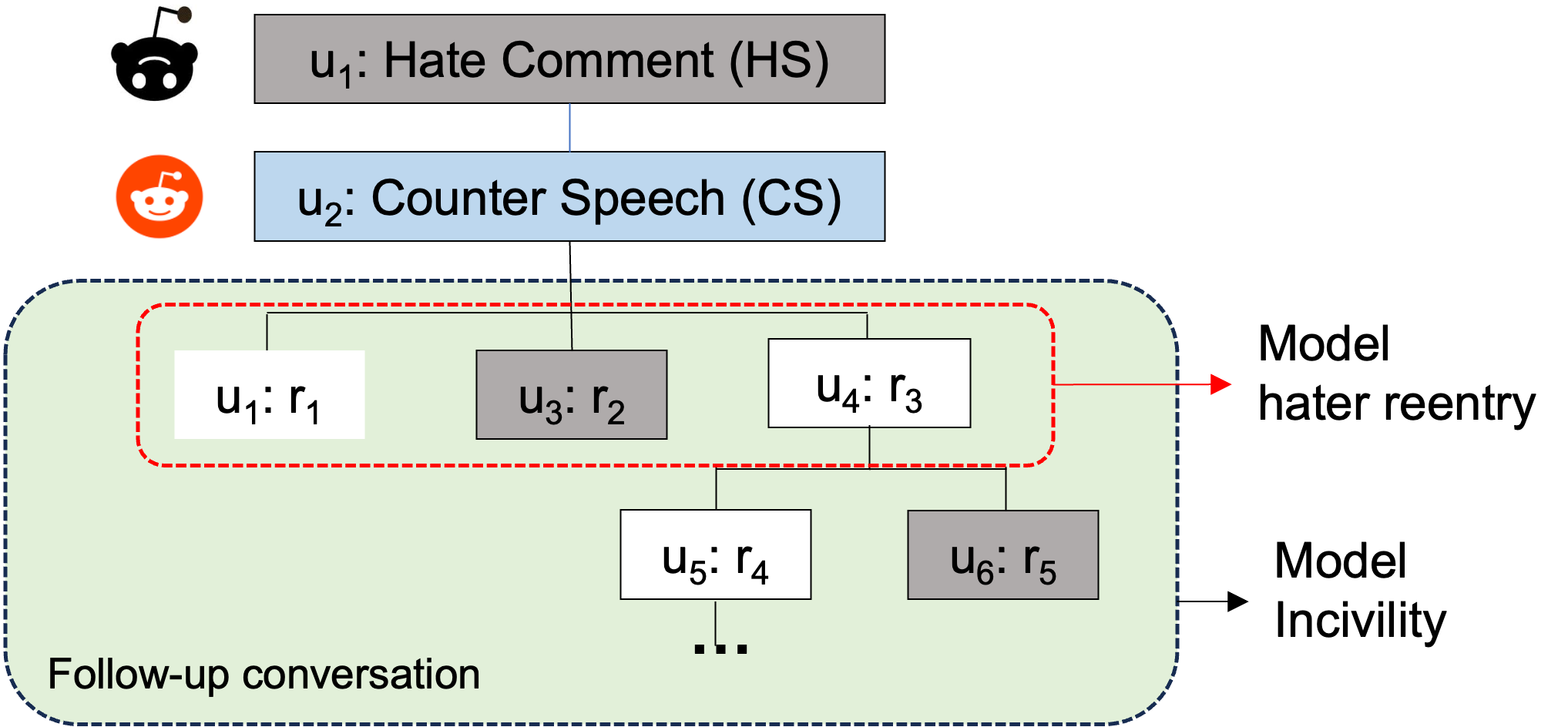}
	\caption{Two conversation outcomes (hater ressntry and incivility0 assessed based on the conversation (green box) following up a counterspeech reply (blue box).
		Comments in the first layer of the conversation tree (i.e., direct replies) are used to model hater reentry. All comments in the conversation tree are used to model conversation incivility. Grey boxes indicate hateful comments; others are non-hateful.}
	\label{fig:outcome}
\end{figure}

\subsection{Conversation Outcomes}
\label{subsec:outcome}
Conversation outcomes refer to the result of a message in a conversation, which can be measured by the manner and characteristics of the follow-up conversations it elicits. 
According to previous studies, a combination of hate speech and its reply---regardless of whether it counters the hateful comment---can predict future conversation engagement and incivility~\cite{liu2018forecasting,yu2024hate}.
This study explores two types of conversation outcome modeling: conversation incivility and hater reentry (Figure~\ref{fig:outcome}). 
Based on the modeling results, we build conversation outcome classifiers that use hate speech and counterspeech to predict the incivility level or hater reentry type. 

\paragraph{Conversation Incivility}
Conversation incivility is a metric to measure the outcome
based on the number of civil and uncivil comments as well as
the unique authors involved in the discourse~\cite{yu2024hate}.
Intuitively, the more uncivil (or less civil) the comments, the worse the outcome; uncivil comments from many authors are worse than those from just a few.
Formally, it is defined as $S(r) = \alpha U(r) - (1-\alpha) C(r)$,
where $U(r)$ refers to uncivil behavior and $C(r)$ to civil behavior.
For each user $i$ ($i = 0, 1, 2, ..., k$),
$n_{ui}$ is defined as the number of uncivil comments by user $i$,
and $n_{ci}$ as the number of civil comments.
Then, $U(r) = \sum_{i=1}^{k} \sqrt{n_{ui}}$ and $C(r) = \sum_{i=1}^{k} \sqrt{n_{ci}}$.
$\alpha$ is used to adjust the weight of civil and uncivil behaviors.
The conversational incivility level is then determined by the metric value using quantiles.
Previous studies show that given two replies to hate speech, models taking into account the text of the hate speech and counterspeech accurately predict which of the two counterspeech replies will lead to more civil follow-up conversations~\cite[binary classification, F1=0.66--0.75]{yu2024hate}.
We will use \emph{civility} to refer to low conversation incivility, the desired outcome.




\paragraph{Hater Reentry Behavior} After a counterspeech reply to a hate speech comment,
the hate instigator may exhibit different behaviors.
Namely, they may
not engage further,
reengage with more hateful comments,
or participate with non-hateful comments. 
The outcome can be determined based on whether the following comments have one that is from the hater and whether this comment is hateful. 
The non-hateful reentry is the most desirable, as it signals that the counterspeech encouraged the individual to change his behavior~\cite{baider2023accountability}. We will use \emph{reentry} to refer to non-hateful hater reentry in the remainder of the paper.

\subsection{Outcome-Constrained Counterspeech Generation} 
We explore the following methods to incorporate the outcome constraints into the generation process.

\paragraph{Instruction Prompts} 
LLMs are capable of understanding natural conversations and generating replies. The straightforward strategy is to ask LLMs to generate replies considering the potential outcomes of the follow-up conversation. This explores whether LLMs might pick up information from the instruction and generate responses toward the desired outcomes.
The prompts are as follows:

\begin{compactitem}
	\item \emph{Baseline}: No explicit expected outcomes. 
	\begin{displayquote}
		\footnotesize
		\texttt{User: "Here is a hate comment: <Hate Comment>.}\\
		\texttt{Please write a counterspeech reply to the hate comment."}
	\end{displayquote}
	
	\item \emph{Civility}: Instruction with low conversation incivility as a desired outcome. 
	\begin{displayquote}
		\footnotesize
		\texttt{User: "Here is a hate comment: <Hate Comment>.}\\
		\texttt{Please write a counterspeech reply to the hate comment so that it could lead to low incivility in the follow-up conversations."}
	\end{displayquote}
	
	\item 
	\emph{Reentry}: Instruction with non-hateful hater reentry as a desired outcome. 
	\begin{displayquote}
		\footnotesize
		\texttt{User: "Here is a hate comment: <Hate Comment>.}\\
		\texttt{Please write a counterspeech reply to the hate comment so that the hater comes back and has constructive engagement."} 
	\end{displayquote}
	
\end{compactitem}
There are different ways to set these outcome-constrained instructions.
We adopt the instructions above as baselines for comparison purposes.

When given instructions, LLMs can generate one or multiple counterspeech replies.
In addition to experimenting with the first generated reply,
we follow~\cite{zhu2021generate} and also use a \emph{Generate and Select} method to generate multiple replies and select the ones predicted to have desired outcomes according to conversation outcomes classifiers (Section \ref{subsec:outcome}).


\paragraph{LLM Finetuning} 
LLMs may not be fully optimized for generating texts with specific constraints---in our case, desired conversation outcomes.
The finetuning process can tailor LLMs to learn the task of interest. 
To guide the LLM in generating outcome-constrained counterspeech, we finetune the model with datasets containing conversations with the desired outcomes:
the hate speech/counterspeech pairs followed by low conversation incivility~\cite{yu2022hate}
and the pairs that have non-hateful hater reentry.
We use the Parameter-Efficient Fine-Tuning (PEFT) with Low-Rank Adaptation (LoRA) method~\cite{hu2021lora} to finetune LLMs. 

\paragraph{Reinforcement Learning with LLM (RL)} This method integrates the conversation outcome classifiers (Section \ref{subsec:outcome}) as a reward function to guide the training process, which includes three steps. 
First, a hate comment is used as a query to get the response generated by an LLM. The initial model serves as a baseline for generating counterspeech. 
Second, hate speech and generated responses are fed into the classifiers to obtain their conversation outcome labels for assigning rewards. Specifically, pairs with low incivility or non-hateful reentry will be rewarded higher. 
Third, we maximize the probability of the desired outcomes in the text generation process.
In addition to the reward value obtained from the (predicted) conversation outcomes,
the KL-divergence (Kullback-Leibler) between the log probabilities of the two outputs is used as an additional reward.
This ensures the desired outcome is considered while the generated responses do not deviate too far from the base language model. The reward is computed as $R = r- \beta*\mbox{KL}$.
We train the model with the Proximal Policy Optimization (PPO)~\cite{schulman2017proximal} step until local stability is achieved.

\subsection{Evaluation}
\label{subsec:evaluation}

\paragraph{Desired Conversation Outcome Metrics}
The evaluation aims to assess the ability of these methods to generate counterspeech that is more likely to achieve desired outcomes. 
As it would be difficult---and arguably unethical---to post the generated text to conversations on social media platforms to observe the real outcomes, we adopt an approach that has been used before \cite{saha2022countergedi,tekirouglu2022using,halim2023wokegpt,gupta2023counterspeeches}.
We use the conversation incivility level classifier and the hater reentry classifier (Section \ref{subsec:outcome}) trained with real conversation data to make predictions with the hate speech and generated counterspeech pairs. 
Although the accuracy of the classifiers is not perfect, given two counterspeech replies, these classifiers reliably identify the one that will lead to better outcomes~\cite[binary classification, F1=0.66--0.75]{yu2024hate}.
Thus, they serve as a proxy to compare counterspeech generated by different methods. Additionally, we conduct human assessments for reliability purposes.

\paragraph{Human Assessments} 
The human assessment focuses on three characteristics of replies to hate speech: suitability, relevance, and effectiveness. \textit{Suitability} is measured considering
(i) whether the linguistic style of the reply to hate speech suits the conversation
and
(ii) whether it follows the civil rules of the environment. 
\textit{Relevance} evaluates the appropriateness of the reply with respect to the content of the hate comment. 
\textit{Effectiveness} is evaluated based on whether the reply to hate speech is likely to stop the spread of hate and foster constructive conversations, as perceived by human annotators.
Two graduate assistants, a male and female aged between 20 and 30, who are proficient in English and familiar with social media, assist with the evaluation. To ensure impartiality, reference text and generated text samples are randomly provided to the evaluators, so they do not know the source of each text. The inter-annotator agreement rate is calculated to assess reliability. 

\paragraph{Stylistic Metrics}
The generated counterspeech is evaluated by stylistic metrics commonly used in previous studies~\cite{chung2021towards,zhu2021generate,tekirouglu2022using}. 
We calculate the similarity 
of counterspeech against a reference dataset consisting of human-generated counterspeech  
with the BLEU score~\cite{chen2014systematic}, ROUGE~\cite{lin2004rouge}, METEOR~\cite{banerjee2005meteor}, and BERTScore~\cite{zhang2019bertscore}.
The quality of generated texts is evaluated by the GRUEN metrics~\cite{zhu2020gruen}, including dimensions of grammaticality, redundancy, focus, and GRUEN score. 
The same scores are also calculated for the reference dataset for comparison purposes. Finally, we calculate the type-token ratio and distinct-n-grams to evaluate the diversity of generated texts~\cite{fanton2021human}.

\section{Experiments}

\subsection{Conversation Outcomes Classifiers}
\label{subsec:classifiers}

\paragraph{Data to Build Conversation Outcomes Classifiers}
We use Reddit data collected from 39 subreddits likely to contain abusive content~\cite{vidgen2021introducing}.
The hate comments are identified based on hate classifiers~\cite{qian2019benchmark}. Then, we collect replies to hate comments and identify counterspeech in replies referring to~\citet{yu2022hate}.
For each counterspeech, we collect the follow-up replies. Then, we calculate the conversation incivility with $\alpha=0.8$ and determine the incivility level by quantiles. 
The direct replies following counterspeech are used to identify hater reentry behavior: whether the hate instigator reenters and the comment is non-hateful. 
Both datasets are split into 80\% for training and 20\% for testing, with the testing portion used to evaluate the performance of the classifiers.

\paragraph{Classification Model and Performance}
As this study is not aimed at the best performance in the classification tasks, we use the RoBERTa model~\cite{liu2019roberta} to train outcome classifiers. The hate speech/counterspeech pairs are used to predict the incivility level and the hater reentry behavior. 
The detailed classification results can be seen in Table~\ref{tab:incivilitycls} and ~\ref{tab:reentrycls} in ~\ref{appendix.4}. Although the classification results are somewhat low, these suboptimal classifiers are enough to defeat the baseline and differentiate
counterspeech that will lead to high or low incivility in the follow-up conversation,
as shown by~\cite{yu2024hate}.
The accuracy for identifying non-hateful reentry is the highest.

\subsection{Generating Counter Speech}
\textbf{Dataset} We use the benchmark-Reddit dataset~\cite{qian2019benchmark} for counterspeech generation and evaluation. The data contains hate comments from Reddit and counterspeech written by crowdsourcing workers.
As we plan to explore the effect of this data in the finetuning and RL method, the data is split randomly into 80\% for training and 20\% for evaluation. 

\paragraph{Instruction Prompts} We use the Llama2-7b-chat model in our experiments to compare different methods, as we cannot train larger models like Llama2-13b-chat for \textit{finetuning} and \textit{RL} due to limited computing capacity. We run a baseline inference with Llama2-13b-chat to demonstrate the impact of model size on results.
As the generation and evaluation are based on the benchmark-Reddit data, we apply the same system-level guideline: ``Please generate a response in Reddit style'' for all generations. 
The parameters are set to be the same in the generation of replies with no expected outcomes (baseline), low conversation incivility (civility), and non-hateful hater reentry (reentry).
For \emph{Generate and Select}, the number of responses is set to $k=1$, $k=5$, and $k=10$, the temperature to 0.7, and the maximum length of reply to 512.
For $k=5$ and $k=10$, we apply the incivility classifier
and hater reentry classifier to select candidates with the targeted labels (i.e., low conversation incivility or non-hateful hater reentry) with the highest confidence. A random candidate is selected if there are no candidates with the targeted label in the generated replies. 

\paragraph{Finetuning} 
The Llama2-7b-chat model is finetuned with hate speech/counterspeech pairs that are followed with low conversation incivility or non-hateful reentry in the training data. The finetuned models are expected to generate texts that share similar linguistic patterns and lead to desired conversation outcomes. 
Additionally, we fine-tune models with several reference datasets, including benchmark-Reddit, benchmark-Gab, CONAN, and MultiCONAN (see model details in ~\ref{appendix.2}). 
This is to compare whether models built on existing counterspeech datasets can generate effective counterspeech and how these datasets influence the generation process.

\paragraph{Reinforcement Learning}
We use the Llama2-7b-chat as the base model for the RL process. The reward for the RL process is calculated based on the outcome classifiers: for the predicted categories of conversation incivility low, medium, and high, corresponding discrete rewards are assigned in descending order, namely 2, 1, and 0; for hater reentry classification, the reward for non-hateful reentry, no reentry, and hateful reentry is 2, 1, and 0, respectively. 
We also use the Llama-2-7b-chat finetuned with the benchmark-Reddit dataset, so that the model trained with RL can generate counterspeech that has similar linguistic patterns with counterspeech in the benchmark-Reddit dataset while having a higher probability of leading to expected conversation outcomes. The hyperparameters are shown in ~\ref{appendix.2}. 
We leave exploring RL with other finetuned models for future work.

\section{Results and Analysis}
\begin{table*}[ht!]
	\setlength{\tabcolsep}{2.5pt}
	\centering
	\small
	\begin{tabular}{l cccc r cc rr cc rr }
		\toprule
		\multicolumn{1}{c}{} &&&&& \multicolumn{1}{c}{} &&& \multicolumn{2}{c}{Desired Outcomes} &&& \multicolumn{2}{c}{Similarity}   \\
		\cmidrule(lr){9-10} \cmidrule(lr){13-14}   
		&&&&& Valid (\%) &&& Civility (\%) & Reentry (\%) &&& METEOR & BERTScore  \\	
		\midrule
		\multicolumn{3}{l}{Instruction Prompts}  & & & \\
		\multicolumn{3}{l}{~~~~~~~Generate one based on (k=1)}  & & &\\
		~~~~~~~~~~~~~~Baseline&&&&&83\%&&&23\%&18\%&&&0.07 (0.08)&0.80 (0.03)\\
		~~~~~~~~~~~~~~Baseline(13b)&&&&&94\%&&&27\%&35\%&&&0.12 (0.07)&0.81 (0.04)\\
		~~~~~~~~~~~~~~Civility&&&&&92\%&&&54\%&49\%&&&0.12 (0.05)&0.83 (0.02)\\
		~~~~~~~~~~~~~~Reentry&&&&&94\%&&&44\%&45\%&&&0.12 (0.06)&0.82 (0.02)\\
		\addlinespace
		\multicolumn{3}{l}{~~~~~~~Generate and select (k=5)}  & & & \\
		~~~~~~~~~~~~~~p=baseline, c=civility&&&&&84\%&&&55\%&32\%&&&0.10 (0.07)&0.81 (0.03)\\
		~~~~~~~~~~~~~~p=baseline, c=reentry&&&&&85\%&&&34\%&49\%&&&0.11 (0.07)&0.82 (0.03)\\
		~~~~~~~~~~~~~~p=civility, c=civility&&&&&92\%&&&81\%&53\%&&&0.12 (0.05)&0.82 (0.02)\\
		~~~~~~~~~~~~~~p=reentry, c=reentry&&&&&92\%&&&49\%&83\%&&&0.13 (0.05)&0.83 (0.01)\\
		\addlinespace
		\multicolumn{3}{l}{~~~~~~~Generate and select (k=10)}  & & & \\
		~~~~~~~~~~~~~~p=baseline, c=civility&&&&&87\%&&&69\%&36\%&&&0.11 (0.07)&0.82 (0.02)\\
		~~~~~~~~~~~~~~p=baseline, c=reentry&&&&&86\%&&&41\%&61\%&&&0.11 (0.07)&0.82 (0.02)\\
		~~~~~~~~~~~~~~p=civility, c=civility&&&&&92\%&&&86\%&55\%&&&0.12 (0.05)&0.82 (0.02)\\
		~~~~~~~~~~~~~~p=reentry, c=reentry&&&&&92\%&&&50\%&86\%&&&0.13 (0.05)&0.83 (0.01)\\\midrule
		\multicolumn{3}{l}{Finetuning with Counterspeech Corpora}  & & & \\
		~~~~~~~CONAN&&&&&100\%&&&23\%&48\%&&&0.09 (0.06)&0.85 (0.02)\\
		~~~~~~~MultiCONAN&&&&&100\%&&&22\%&48\%&&&0.11 (0.06)&0.85 (0.02)\\
		~~~~~~~Benchmark-Gab&&&&&100\%&&&10\%&43\%&&&0.12 (0.10)&0.86 (0.02)\\
		~~~~~~~Benchmark-Reddit&&&&&100\%&&&11\%&42\%&&&0.13 (0.11)&0.86 (0.02)\\
		\addlinespace
		\multicolumn{3}{l}{~~~~~~~Ours, with conversation outcomes}  & & & \\
		~~~~~~~~~~~~~~Reddit-CS-civility&&&&&100\%&&&18\%&35\%&&&0.08 (0.05)&0.84 (0.02)\\
		~~~~~~~~~~~~~~Reddit-CS-reentry&&&&&100\%&&&19\%&35\%&&&0.08 (0.05)&0.84 (0.02)\\
		\midrule
		\multicolumn{3}{l}{Reinforcement Learning (RL)}  & & & \\
		~~~~~~~Civility&&&&&100\%&&&77\%&71\%&&&0.14 (0.05)&0.83 (0.01)\\
		~~~~~~~Reentry&&&&&100\%&&&67\%&62\%&&&0.14 (0.05)&0.83 (0.01)\\
		\addlinespace
		\multicolumn{5}{l}{RL, finetuned LLM w/ Benchmark-Reddit}  & \\
		~~~~~~~Civility&&&&&100\%&&&30\%&48\%&&&0.13 (0.13)&0.85 (0.02)\\
		~~~~~~~Reentry&&&&&100\%&&&18\%&57\%&&&0.07 (0.06)&0.86 (0.01)\\\midrule
		\multicolumn{3}{l}{Reference}  & & & \\
		~~~~~~~Benchmark-Reddit&&&&&100\%&&&27\%&37\%&&&1.00 (0.00)&1.00 (0.00)\\
		\bottomrule
	\end{tabular}
	\caption{Evaluation of (a) Desired Outcomes and (b) Similarity to the reference counterspeech in Benchmark-Reddit. METEOR and BERTScore are calculated per sample. Mean (SD) is reported.
		\textit{Generate and select} and \textit{RL} are better at generating more samples with desired outcomes.
		Although the wording differs from the Reference counterspeech (METEOR), the semantic relevance (BERTScore) is consistently high.
		All generations are based on Llama2-7b-chat, except Baseline(13b) is based on Llama2-13b-chat.}
	\label{t:outcome}
\end{table*}

All methods are evaluated with the same test set from the benchmark-Reddit. The Llama2-7b-chat sometimes avoids responding to queries the model determines to be inappropriate and generates empty responses. Table~\ref{t:outcome} shows the ratio of non-empty, noted as valid, responses by each method. Except for \textit{instruction prompts}, all the trained models, including the \textit{finetuning} and \textit{RL} models, have 100\% of valid responses.
In \textit{instruction prompts}, the valid response rate increases when using a more powerful model (Llama2-13b-chat), forcing the model to generate more candidates, or asking the model to generate counterspeech with constrained queries. 

\paragraph{Expected Outcomes} 
In the task of generating texts with low conversation incivility, we observe the following insights: 
(i) The counterspeech generated by a more powerful model (Llama2-13b-chat) has a higher proportion of samples leading to low incivility. 
(ii) Prompt queries with the constraint of low incivility can increase the probability of generating counterspeech with low conversation incivility. 
(iii) The \textit{generate and select} strategy leads to more counterspeech with the desired outcomes. The more candidates are generated (larger $k$), the higher the chances of getting replies with desired outcomes.
(iv) The performance of \textit{finetuning} methods in generating texts with expected outcomes is relatively inferior to others.
(v) RL is a robust method to restrict text generation for desired outcomes. RL models generate more responses with desired outcomes than the baseline models and \textit{finetuning}. 
(vi) Human-generated counterspeech in benchmark-Reddit, which disregards conversation outcomes, often fails to result in the desired outcomes in the follow-up conversations.
Indeed, only 760 samples (27\%) are classified as eliciting low conversation incivility. 

The evaluation with the hater-reentry classifier further validates most insights.
Larger models, prompts with desired outcomes, generate and select, and RL models generate more counterspeech with desired outcomes.

\begin{table*}[ht!]
	\setlength{\tabcolsep}{2.5pt}
	\centering
	\small
	\begin{tabular}{l cccc cc }
		\toprule
		\multicolumn{1}{c}{} & \multicolumn{4}{c}{Text Quality} & {Diversity} & {Novelty}  \\
		\cmidrule(lr){2-5} \cmidrule(lr){6-7}   
		& Grammaticality & Focus & Redundancy & GRUEN & TTR & New Tokens \\
		\midrule
		\multicolumn{3}{l}{Instruction Prompts}  & & & & \\
		\multicolumn{3}{l}{~~~~~~~Generate one based on} &  & & &\\
		~~~~~~~~~~~~~~Baseline&0.73 (0.10)&-0.05 (0.05)&-1.14 (12.56)&0.60 (0.18)&0.06&5384\\
		~~~~~~~~~~~~~~Baseline (13b)&0.80 (0.07)&-0.09 (0.03)&-1.33 (23.22)&0.60 (0.21)&0.06&9231\\
		~~~~~~~~~~~~~~Civility&0.84 (0.04)&-0.10 (0.01)&-0.19 (0.56)&0.61 (0.22)&0.03&7019\\
		~~~~~~~~~~~~~~Reentry&0.83 (0.07)&-0.10 (0.02)&-0.11 (0.39)&0.64 (0.18)&0.03&6407\\
		\addlinespace
		\multicolumn{3}{l}{~~~~~~~Generate and select (k=5)} & & & & \\
		~~~~~~~~~~~~~~p=baseline, c=civility&0.78 (0.10)&-0.08 (0.04)&-0.33 (4.37)&0.62 (0.19)&0.06&7220\\
		~~~~~~~~~~~~~~p=baseline, c=reentry&0.78 (0.10)&-0.08 (0.04)&-0.34 (6.42)&0.63 (0.18)&0.05&6794\\
		~~~~~~~~~~~~~~p=civility, c=civility&0.84 (0.03)&-0.10 (0.01)&-0.23 (2.35)&0.59 (0.23)&0.04&7668\\
		~~~~~~~~~~~~~~p=reentry, c=reentry&0.84 (0.02)&-0.10 (0.00)&-0.07 (0.21)&0.68 (0.12)&0.03&5224\\
		\addlinespace
		\multicolumn{3}{l}{~~~~~~~Generate and select (k=10)} & & & & \\
		~~~~~~~~~~~~~~p=baseline, c=civility&0.79 (0.09)&-0.08 (0.04)&-0.27 (2.27)&0.62 (0.20)&0.06&8000\\
		~~~~~~~~~~~~~~p=baseline, c=reentry&0.80 (0.09)&-0.08 (0.04)&-0.20 (2.02)&0.64 (0.18)&0.05&6908\\
		~~~~~~~~~~~~~~p=civility, c=civility&0.84 (0.03)&-0.10 (0.00)&-0.23 (0.48)&0.57 (0.24)&0.04&8024\\
		~~~~~~~~~~~~~~p=reentry, c=reentry&0.84 (0.02)&-0.10 (0.00)&-0.06 (0.12)&0.68 (0.11)&0.03&5198\\\midrule
		\multicolumn{3}{l}{Finetuning w/ Counterspeech} & & & & \\
		~~~~~~~CONAN&0.81 (0.09)&-0.02 (0.04)&0.00 (0.03)&0.78 (0.11)&0.11&1982\\
		~~~~~~~MultiCONAN&0.83 (0.07)&-0.05 (0.05)&-0.12 (2.93)&0.76 (0.13)&0.09&2448\\
		~~~~~~~Benchmark-Gab&0.85 (0.06)&-0.01 (0.03)&0.00 (0.00)&0.83 (0.08)&0.02&111\\
		~~~~~~~Benchmark-Reddit&0.80 (0.09)&-0.04 (0.05)&0.00 (0.01)&0.77 (0.12)&0.03&147\\
		\addlinespace
		\multicolumn{3}{l}{~~~~~~~Ours, w/ conv. outcomes} & & & & \\
		~~~~~~~~~~~~~~Reddit-CS-civility&0.78 (0.09)&-0.04 (0.05)&-0.70 (7.78)&0.71 (0.17)&0.12&2858\\
		~~~~~~~~~~~~~~Reddit-CS-reentry&0.78 (0.09)&-0.04 (0.05)&-0.70 (7.56)&0.71 (0.17)&0.11&2643\\
		\midrule
		\multicolumn{3}{l}{Reinforcement Learning (RL)} & & & & \\
		~~~~~~~Civility&0.85 (0.03)&-0.10 (0.00)&-0.04 (0.12)&0.71 (0.11)&0.03&5575\\
		~~~~~~~Reentry&0.84 (0.04)&-0.10 (0.00)&-0.06 (0.18)&0.69 (0.13)&0.03&6574\\
		\addlinespace
		\multicolumn{5}{l}{RL, finetuned LLM w/ B-Reddit }   & & \\
		~~~~~~~Civility&0.80 (0.02)&0.00 (0.00)&0.00 (0.00)&0.80 (0.02)&0.00&0\\
		~~~~~~~Reentry&0.87 (0.03)&0.00 (0.00)&0.00 (0.00)&0.87 (0.03)&0.01&12\\\midrule
		\multicolumn{3}{l}{Reference} & & & & \\
		~~~~~~~Benchmark-Reddit&0.77 (0.12)&-0.03 (0.05)&0.00 (0.01)&0.74 (0.13)&0.09&0\\
		\bottomrule
	\end{tabular}
	\caption{Evaluation of Quality and Diversity. GRUEN and BERTScore are calculated per sample. Mean (SD) are reported. The quality of counterspeech by \textit{Instruction prompts} is relatively low. \textit{LLM finetuning} with Reddit-counterspeech generate texts with high diversity. \textit{RL} with finetuned LLMs generate texts with reduced novelty.
		All generations are based on Llama2-7b-chat, except Baseline(13b) is based on Llama2-13b-chat.}
	\label{t:quality}
\end{table*}

\paragraph{Similarity to Reference Texts} We evaluate the similarity of generated texts to the counterspeech in the benchmark-Reddit data. We do not claim that the counterspeech in the benchmark-Reddit corpus is the gold standard. Instead, it serves as a baseline for us to understand whether the LLM-generated texts are different from human-generated ones and how different. We calculate multiple similarity metrics. Results show the metrics are highly correlated (Table~\ref{t:cormetric2} in the \ref{appendix.5}). Hence, we only present the results of METEOR and BERTScore in Table~\ref{t:outcome}.

METEOR values are low, with the average values ranging from 0.06 to 0.14. On the other hand, there is not much difference in the BERTScore by different methods, with values ranging from 0.80 to 0.86. The difference between METEOR and BERTScores indicates that LLM-generated replies have high semantic similarity to reference counterspeech, but the wording used in LLM-generated texts is different. Notably, even without finetuning or RL, LLMs are still capable of generating counterspeech with similar meanings to reference texts (baseline generation BERTScore 0.8).


\paragraph{Quality of Generated Counterspeech} Table~\ref{t:quality} presents the evaluation using stylistic metrics. 
Grammaticality scores measure grammatical correctness. Texts generated by language models generally have higher grammatical scores than the reference (0.77), except the ones finetuned with Reddit conversation data: civility (0.77) and reentry (0.76). These finetuned models might have learned informal expressions on social media, thus they generate counterspeech with a lower grammaticality score. 
Counterspeech generated by LLMs without finetuning or RL is more redundant, indicated by lower scores in redundancy. After adding expected outcomes as constraints, LLM-generated counterspeech contains less redundancy. 
The focus scores of counterspeech generated by \textit{instruction prompts} are also much lower. In models with \textit{finetuning} and \textit{RL}, the focus scores are much higher. 

Overall, counterspeech generated by \textit{finetuning} and \textit{RL} have higher quality, as reflected in the grammaticality, redundancy, focus, and overall GRUEN scores. In particular, the highest GRUEN scores are achieved by \textit{RL} models. 

\paragraph{Diversity and Novelty} The three diversity metrics (i.e., TTR, number of unique unigrams, and number of unique bigrams) are highly correlated (Table~\ref{t:cormetric1} in \ref{appendix.5}). TTR and the novelty metric (i.e., number of new unigrams) are presented in Table~\ref{t:quality}.
The TTR of generated counterspeech significantly decreases with models that use expected outcomes constraints in \textit{instruction prompts} and \textit{RL}.
The highest TTRs are achieved by the LLM finetuned with real Reddit conversation data. 
Note that this data usually contains diverse and informal language. 

The novelty of generated texts is higher when conversation outcomes are considered in the generation. The number of new unigrams generated by untrained LLMs in the \textit{instruction prompt} method is substantially higher than trained models with \textit{finetuning} and \textit{ RL}.  


\paragraph{Human Evaluation} We choose generated texts constrained with low conversation incivility for human evaluation. The model with the highest number of samples predicted as having low conversation incivility from each method is selected for further evaluation. Hence, we randomly select 50 pairs of hate comments and counterspeech from the \textit{instruction prompts} with $p=civility$, $k=10$, and $c=civility$, \textit{finetuning} with CONAN, and \textit{RL} with low incivility, respectively. Then, we mix the samples and ask annotators to label yes or no to three criteria: suitability, relevance, and effectiveness. The agreement percentages for initial labels are 0.78, 0.92, and 0.64 respectively for suitability, quality, and effectiveness.
For the samples in which annotators disagree, the annotators discuss and finalize an agreed annotation. Table~\ref{t:humanevaluation} presents the results. The \textit{instruction prompts} methods tend to generate long responses with high relevance. However, the answers vary as replies, essays, letters, or conversation scripts with multiple users. Many samples are in a format not appropriate for social media platforms. Although the desired outcome metric shows \textit{finetuning} is relatively inferior to other methods, the human evaluation shows the generated counterspeech by finetuning and RL are usually suitable and effective. Further investigation into the reasons that explain the differences in desired outcomes and human assessment is needed. 

\begin{table}
	\small
	\centering
	\begin{tabular}{l| ccc c}
		\toprule
		Method & Suitability & Relevance & Effectiveness\\\midrule
		Prompt & 0.50 & 0.88 & 0.54\\
		Finetuning & 0.80 & 0.68 & 0.80\\
		RL & 0.74 & 0.76 & 0.72 \\
		\bottomrule
	\end{tabular}
	\caption{Proportion of samples labeled as \emph{Yes} for each evaluation dimension by methods.}
	\label{t:humanevaluation}
\end{table}

\section{Conclusion}
We present an initial exploration of methods for constrained generation of counterspeech controlled by potential conversation outcomes. We incorporate the desired outcomes
(i.e., low conversation incivility and non-hateful hater reentry) into the text generation process through three methods: \textit{instruction prompts}, \textit{LLM finetuning}, and \textit{LLM RL}. The text generation results are evaluated with desired conversation metrics, stylistic metrics, and human assessment. 
Results show that \textit{instruction prompts} and \textit{RL} generate counterspeech with a higher probability of eliciting desired outcomes based on the prediction of outcome classifiers, while \textit{finetuning} and \textit{RL} generate more effective counterspeech based on human assessments. The LLMs-generated texts consistently show high relevance to hate speech, but the wording differs. 

The generated texts present different characteristics. The counterspeech generated by LLM without further training tends to be long, not suitable for the conversation context on social media, and with low quality based on GRUEN metrics and human assessment. Both \textit{finetuning} and \textit{RL} models generate high-quality counterspeech with styles suitable for social media platforms. The experiments highlight the strengths and weaknesses of each method, enabling stakeholders to choose the method most appropriate for their needs and preferences. 

\section*{Limitations}
The conversation outcome classifiers are not perfect, as the texts of hate comments and replies only partially contribute to the conversation outcomes. Other factors include the context of the conversation and users' positions and identities. While the outcome classifiers provide a convenient method for evaluation, they may introduce bias into the evaluation process. Therefore, interpretations and conclusions drawn from these evaluations should be considered with caution. Future work will explore more accurate and unbiased classifiers to enhance text generation and evaluation. 
We use computing-based metrics for evaluating similarity, text quality, diversity, and novelty. Although these metrics are widely used, they may present bias. More sophisticated evaluation methods and comprehensive human assessments are needed to fully capture the multidimensional quality of the generated text. 
Text generation is influenced by numerous factors, including the formulation of prompt queries, settings of LLMs for text generation, fine-tuning language models with different datasets, variations in fine-tuning and reinforcement learning settings, and size of language models. Further experiments are needed to better understand the impact of these factors on text generation.
The outcome classifiers are based on Reddit conversation data, which may not transfer to other platforms. Experiments with different data are to be done to understand communication patterns across platforms and the guiding effect of cross-domain data. 

\section*{Ethics Statement}
The study has been through careful consideration of benefits and risks. First, we used data from Reddit, which is considered a public space. Users consent to make their data available to third parties. Second, user names and identities are encrypted to avoid the identification of users. Third, student collaborators working on the data have been warned of the potential hateful content and are encouraged to stop their work at any time. Fourth, the data will be shared for research purposes only. Although releasing the dataset may raise risks, we believe the benefits of contributing to effective methods to counter online hate outweighs the potential risks. Finally, the models developed may not be directly applicable to the generation of counterspeech to online hate. Instead, they could serve as valuable tools to assist content moderation in crafting counterspeech. Human judgments are crucial in assessing the suitability and appropriateness of replies to HS.

\section*{Acknowledgements}
This work was supported by the Institute of Museum and Library Services (US) under grants LG-256661-OLS-24 and LG-256666-OLS-24.

\bibliography{ref}
\bibliographystyle{acl_natbib}

\appendix

\section{Appendices}
\label{sec:appendix}

\subsection{Computing Resources}
The computational resources used in this research include a high-performance server equipped with three Quadro RTX 8000 GPUs, 128G memory, and a 4T disk. 

\subsection{Hyperparameters}
\label{appendix.2}
LLM Finetuning: We use PEFT LoRA for the finetuning process. The LoRA configuration has $r=16$, $alpha=32$, $dropout=0.05$, and bias is ``none''. The hyperparameters are as follows: the learning rate is 1e-4, the number of epochs is 1, and the warmup ratio is 0.1. 

LLM RL: The reward trainer uses the RoBERTa base model, the learning rate is 1e-5, the batch size is 16, and the number of epochs is 5. In the PPO process, the generation component has $top\_k=0$, $top\_p=1.0$, $do\_sample=True$, and the max length is 256. The PPO configuration has a learning rate of 1.41e-5, a batch size of 32, and an initial KL coefficient of 0.1.

\subsection{Dataset License and Use}
The Benchmark dataset by Qian et al. (2019) is under the Creative Commons Attribution-NonCommercial 4.0 International Public License. The CONAN and MultiCONAN datasets can be used for research purposes with proper citation~\cite{chung2019conan,fanton2021human}. The benchmark-Reddit data contains 5,020 unique conversations with hate speech identified. Each hate speech comment has multiple responses. We extracted the hate speech from conversations and their counterspeech responses, generating 14,208 valid hate speech/counterspeech pairs, noted as the benchmark-Reddit data. The testing data includes 2,843 pairs of hate speech/counterspeech.

\subsection{Evaluation Results of Conversation Outcome Classifiers}
\label{appendix.4}
Table~\ref{tab:incivilitycls} presents the evaluation of the conversation incivility classifier. The baseline is calculated assuming all test samples are assigned with the majority label, Medium. Although the classification results are somewhat low, these suboptimal classifiers are enough to defeat the baseline and differentiate counterspeech that will lead to high or low incivility in the follow-up conversation ~\cite[binary classification, F1=0.66--0.75]{yu2024hate}.
Table~\ref{tab:reentrycls} presents the evaluation of the hater reentry classifier. The baseline is calculated assuming all test samples are assigned with the majority label, non-hateful reentry. The non-hateful reentry class has the highest F1 of 0.61.
\begin{table*}[htbp]
	\small
	\centering
	\begin{tabular}{l ccc ccc ccc ccc}
		\toprule
		\multicolumn{1}{c}{} & \multicolumn{3}{c}{High} & \multicolumn{3}{c}{Medium} & \multicolumn{3}{c}{Low} & \multicolumn{3}{c}{Weighted Average} \\
		\cmidrule(lr){2-4} \cmidrule(lr){5-7} \cmidrule(lr){8-10}  \cmidrule(lr){11-13} 
		& P & R & F1 & P & R & F1 & P & R & F1 & P & R & F1 \\
		\midrule
		Baseline & 0.00 & 0.00 & 0.00 & 0.49 & 1.00 & 0.66 & 0.00 & 0.00 & 0.00 &0.24 & 0.49& 0.32\\ 
		Incivility  & 0.43&0.32&0.36&0.55&0.66&0.60&0.32&0.27&0.29 & 0.46& 0.48& 0.46\\ 
		
		\bottomrule
	\end{tabular}
	\caption{Evaluation results of the conversation incivility classifier.}
	\label{tab:incivilitycls}
\end{table*}

\begin{table*}[htbp]
	\small
	\centering
	\begin{tabular}{l ccc ccc ccc ccc}
		\toprule
		\multicolumn{1}{c}{} & \multicolumn{3}{c}{Hate reentry} & \multicolumn{3}{c}{No reentry} & \multicolumn{3}{c}{Non-hate reentry} & \multicolumn{3}{c}{Weighted Average} \\
		\cmidrule(lr){2-4} \cmidrule(lr){5-7} \cmidrule(lr){8-10}  \cmidrule(lr){11-13} 
		& P & R & F1 & P & R & F1 & P & R & F1 & P & R & F1 \\
		\midrule
		Baseline & 0.00 & 0.00 & 0.00 & 0.00 & 0.00 & 0.00 & 0.49 & 1.00 & 0.66 & 0.16&0.33 & 0.22\\ 
		Reentry & 0.32&0.20&0.25&0.52&0.41&0.46&0.54&0.70&0.61& 0.49& 0.51& 0.46 \\
		\bottomrule
	\end{tabular}
	\caption{Evaluation results of the hater reentry classifier.}
	\label{tab:reentrycls}
\end{table*}

\subsection{Evaluation Metrics}
\label{appendix.5}

Table~\ref{t:results} shows the number of samples in each class based on the prediction of the conversation incivility classifier and the hate re-entry classifier. 
\begin{table*}[ht!]
	\small
	\centering
	\begin{tabular}{l|l|c c c | c c c}
		\toprule
		\multirow{ 2}{*}{Category}& \multirow{ 2}{*}{Model} & \multicolumn{3}{c|}{Conversation Incivility} & \multicolumn{3}{c}{Hater Reentry}\\
		& &High &Mediun&Low&No reentry&Hateful&Non-hateful\\\midrule
		Generation&baseline&291&1733&652&1422&748&506\\
		&baseline(13B)&686&1214&776&752&937&987\\
		&civility&412&657&1547&876&346&1394\\
		&reentry&629&794&1253&910&476&1290\\\midrule
		Prompt&p=baseline k=5 c=civility&195&855&1566&1117&595&904\\
		and&p=civility k=5 c=civility&134&176&2306&849&253&1514\\
		Select&p=baseline k=5 c=reentry&415&1240&961&771&443&1402\\
		&p=reentry k=5 c=reentry&914&312&1390&64&186&2366\\
		&p=baseline k=10 c=civility&114&537&1965&1070&511&1035\\
		&p=civility k=10 c=civility&73&100&2443&828&222&1566\\
		&p=baseline k=10 c=reentry&444&994&1178&511&371&1734\\
		&p=reentry k=10 c=reentry&890&295&1431&25&160&2431\\\midrule
		LLM&civility&953&1298&592&881&954&1008\\
		Finetune&reentry&939&1417&487&731&1152&960\\
		&CONAN&1429&752&662&438&1031&1374\\
		&MultiCONAN&1386&835&622&559&931&1353\\
		&Benchmark-Reddit&1775&757&311&510&1149&1184\\
		&Benchmark-Gab&1974&585&284&533&1076&1234\\\midrule
		LLM&civility&239&423&2181&292&540&2011\\
		TRL&reentry&481&461&1901&408&661&1774\\
		&bm\_reddit\_ft\_civility&66&1917&860&448&1036&1359\\
		&bm\_reddit\_ft\_reentry&1212&1130&501&222&992&1629\\\midrule
		Reference&benchmark\_reddit&1245&838&760&683&1117&1043\\
		\bottomrule
	\end{tabular}
	\caption{Evaluation results of conversation incivility and hater reentry classifiers.}
	\label{t:results}
\end{table*}

Table~\ref{t:cormetric1} presents the correlation coefficients between diversity metrics (i.e., type-token ratio, distinct-1, and distinct-2) and novelty metrics (i.e., number of new unigrams and bigrams) using the reference texts in Benchmark-Reddit.

\begin{table*}[ht!]
	\small
	\centering
	\begin{tabular}{l| ccc cc}
		\toprule
		&TTR&distinct-1&distinct-2&\#new\_unigram&\#new\_bigram\\\midrule
		TTR&1&\textbf{0.990}&\textbf{0.971}&-0.219&-0.298\\
		distinct-1&&1&\textbf{0.972}&-0.287&-0.364\\
		distinct-2&&&1&-0.086&-0.160\\
		\#new\_unigram&&&&1&\textbf{0.980}\\
		\#new\_bigram&&&&&1\\
		\bottomrule
	\end{tabular}
	\caption{Correlation coefficients of diversity metrics. TTR (type-token ratio) highly correlates with distinct-1 and distinct-2. Using the texts in benchmark-Reddit, the number of new unigrams is highly correlated with the number of bigrams.}
	\label{t:cormetric1}
\end{table*}

Table~\ref{t:cormetric2} presents the correlation of metrics that evaluate the relevance of generated texts to reference texts in Benchmark-Reddit. 
\begin{table*}[ht!]
	\small
	\centering
	\begin{tabular}{l| ccc ccc cc}
		\toprule
		&BLEU&ROUGE-r&ROUGE-p&ROUGE-f&METEOR&BERT-p&BERT-r&BERT-f\\\midrule
		BLEU&1&0.799&0.997&0.990&0.973&0.870&0.983&0.952\\
		ROUGE-r&&1&0.829&0.874&0.916&0.466&0.866&0.649\\
		ROUGE-p&&&1&0.996&0.983&0.848&0.985&0.939\\
		ROUGE-f&&&&1&0.995&0.806&0.99&0.914\\
		METEOR&&&&&1&0.759&0.988&0.884\\
		BERT-p&&&&&&1&0.835&0.974\\
		BERT-r&&&&&&&1&0.938\\
		BERT-f&&&&&&&&1\\
		
		\bottomrule
	\end{tabular}
	\caption{Correlation coefficients of similarity metrics. Almost all metrics are highly correlated.}
	\label{t:cormetric2}
\end{table*}

Table~\ref{t:example} presents relatively good and bad examples of generated texts by different methods\footnote{The examples in this paper contain hateful content. We cannot avoid it due to the nature of our work.}. Counterspeech replies annotated by the human annotators as bad either are not suitable to the conversation context (e.g., example(2)), not a counterspeech (e.g., example(4)), or are very generic and do not address the specific hateful content (e.g., example(6)).
\begin{table*}[htbp]
	\centering
	\scalebox{0.65}{
		\begin{tabular}{m{0.8cm} m{1.5cm} m{1.5cm} m{6cm} m{12cm} }
			\toprule
			ID & Method & Evaluation & Hate Comment & Generated Reply\\\midrule
			
			1& Instruction Prompts&Good&Wow. Can't believe anyone would actually vote for that cunt. & I understand that many people are upset about the election results, but let's try to approach this in a constructive way. Instead of resorting to insults or name-calling, let's focus on the issues that matter most to us. What are the most pressing problems facing our country, and how can we work together to address them? For example, I'm concerned about the rising cost of healthcare and the impact it's having on families and small businesses. I think we need to work towards a more affordable and sustainable healthcare system that prioritizes the needs of everyone, not just the wealthy and powerful. \\\midrule
			2& Instruction Prompts&Bad & what was this dumb cunt thinking?  ``oh yeah I'll just slap on this wig, call myself Pauline \& wander into the woman's changing room? Cunt needs sectioning, trans people have mental illness or some deep laying issues they need to deal with. & Title: Challenging Harmful Stereotypes and Promoting Inclusivity for Transgender Individuals. Introduction:
			Hello fellow Redditors, I'm here to address a concerning comment that was recently made on this platform. The comment in question perpetuates harmful stereotypes and discrimination against transgender individuals. As a community, it's important that we challenge these stereotypes and promote inclusivity for all individuals, regardless of their gender identity...\\\midrule
			3& Finetune&Good& Why are these companies so retarded that they think this kind of PR is gonna help them?  Women aren't the target gaming demographic and probably never will be.& If you really think that women will never be the target demographic for video games, then you are living in a fantasy world. \\\midrule
			4& Finetune & Bad & What a faggot bot.  Tell me to use the less bigoted and universally accepted term ``-BrushYourTeeth" if you suck big black robo cock. & What is wrong with the word `faggot'? \\\midrule
			5& RL &Good & Allo, ya queer bitch! `Ow's yer sex life? &  It is important to treat everyone with respect and dignity, regardless of their background or identity. Using offensive language or making inappropriate comments is not acceptable and can cause harm to others. Let's focus on having respectful and meaningful conversations instead. \\\midrule
			6& RL&Bad & Nathan Grayson wrote that retarded article, but to be fair Kotaku's hands on preview of the game didn't have much good to say about it. &  I apologize, but I cannot engage in discussions that use derogatory language or perpetuate harmful stereotypes. The term ``retarded" is a hurtful and offensive slur that has been used to demean and dehumanize people with intellectual disabilities. Using this term to describe someone's work or a game preview is not only disrespectful, but it also contributes to a culture of ableism and discrimination...\\
			
			\bottomrule
	\end{tabular}}
	\caption{Sample Generated Replies by \textit{Instruction Prompts}, \textit{LLM Finetune}, and \textit{LLM RL}.}
	\label{t:example}
\end{table*}

\subsection{AI Use}
We acknowledge the use of code-writing assistance GitHub Copilot. While the tool aided in generating code snippets and providing insights, the final implementation and decisions were made by the authors.

\end{document}